\documentclass[runningheads]{llncs}
\usepackage{graphicx}
\usepackage{xcolor}
\usepackage{tikz}
\usepackage{svg}
\usepackage{tabularx}
\usepackage{pgfplots}
\usepackage{float}
\usepackage{booktabs}
\usepackage[utf8]{inputenc}
\usepackage[T1]{fontenc}
\usepackage{combelow}
\usepackage{newunicodechar}
\usepackage{subfiles}
\usepackage{epstopdf}

\makeatletter
\newcommand{\printfnsymbol}[1]{
  \textsuperscript{\@fnsymbol{#1}}
}
\makeatother

\begin{document}

\title{Self-Supervised Representation Learning on Document Images}

\author{Adrian Cosma\inst{1}\inst{2}
Mihai Ghidoveanu\inst{1}\inst{3}\thanks{Equal contribution}
Michael Panaitescu-Liess\inst{1}\inst{3}\printfnsymbol{1} \\
Marius Popescu\inst{1}\inst{3}}

\authorrunning{Cosma et al.}

\institute{Sparktech Software \\
\email{\{adrian.cosma, mihai.ghidoveanu, michael.panaitescu\}@sparktech.ro} \and
University Politehnica of Bucharest \and
Faculty of Mathematics and Computer Science, University of Bucharest \\
\email{popescunmarius@gmail.com}
}

\maketitle
\begin{abstract} 
This work analyses the impact of self-supervised pre-training on document images in the context of document image classification. While previous approaches explore the effect of self-supervision on natural images, we show that patch-based pre-training performs poorly on document images because of their different structural properties and poor intra-sample semantic information. We propose two context-aware alternatives to improve performance on the Tobacco-3482 image classification task. We also propose a novel method for self-supervision, which makes use of the inherent multi-modality of documents (image and text), which performs better than other popular self-supervised methods, including supervised ImageNet pre-training, on document image classification scenarios with a limited amount of data.

\end{abstract}

\section{Introduction}
A document analysis system is an important component in many business applications because it reduces human effort in the extraction and classification of information present in documents. 

While many applications use Optical Character Recognition systems (OCR) to extract text from document images and directly operate on it, documents often have an implicit visual structure. Helpful contextual information is given by the position of text in a page and, generally, the page layout. Reports containing tables and figures, invoices, resumes, and forms are difficult to process without considering the relationship between layout and textual content.
 
As such, while there are efforts in dealing with the visual structure in documents leveraging text \cite{liu-etal-2019-graph,DBLP:journals/corr/YangYAKKG17}, relevant-sized datasets are mostly internal, and privacy concerns inhibit public release. Moreover, labeling of such datasets is an expensive and time-consuming process. 

This is not the case for natural images. Natural images are prevalent on the internet, and large-scale annotated datasets are publicly available. The ImageNet database \cite{imagenet_cvpr09} contains 14M annotated natural images with 1000 classes and has powered many advances in computer vision and image understanding through training of high capacity convolutional neural networks (CNNs). ImageNet also provides neural networks with the ability to transfer the information to other unrelated tasks like object detection and semantic segmentation \cite{DBLP:journals/corr/HuhAE16}. A neural network pre-trained on ImageNet has substantial performance gains compared to a network trained from scratch \cite{zhuang2019comprehensive}.

However, it was shown that pre-training neural networks with large amounts of noisily labeled images \cite{DBLP:journals/corr/abs-1905-00546} substantially improves the performance after fine-tuning on the main classification task. This is indicative of a need to make use of a large corpus of partially labeled or unlabeled data. Moreover, modern methods of leveraging unlabeled data have been developed \cite{DBLP:journals/corr/DoerschGE15,DBLP:journals/corr/NorooziF16}, by creating a pretext task, in which the network is under self-supervision, and afterwords fine-tuning on the main task. 

Unlike natural image datasets, document datasets are hard to come by, especially fully annotated ones, and have only a fraction of the scale of ImageNet \cite{harley2015icdar,li2019tablebank}. However, unlabeled documents are easily found online in the form of e-books and scientific papers.

Qualitatively, document images are very different from natural images, and therefore using a pre-trained CNN on ImageNet for fine-tuning on documents is questionable. Document images are also structurally different from natural images, as they are not invariant to scaling and flips. It has been shown that models trained on ImageNet often generalize poorly to fine-grained classification tasks on classes that are poorly represented in ImageNet \cite{DBLP:journals/corr/abs-1805-08974}. While there are classes that are marginally similar to document images (i.e. menus, websites, envelopes), they are vastly outnumbered by other natural images. Moreover, models that are pre-trained on RVL-CDIP \cite{harley2015icdar} dataset have a much better performance on document classification tasks with a limited amount of data \cite{DBLP:journals/corr/abs-1711-05862}. 

Self-supervision methods designed for document images have received little attention. As such, there is a clear need for learning more robust representations of documents, which make use of large, unlabeled document image datasets.

This paper makes the following contributions to the field of document understanding:

\begin{enumerate}
   \item We make a quantitative analysis of self-supervised methods for pre-training convolutional neural networks on document images.
   
   \item We show that patch-based pre-training is sub-optimal for document images. To that end, we propose improved versions of some of the most popular methods for self-supervision that are better suited for learning structure from documents.
   
   \item We propose an additional self-supervision method which exploits the inherent multi-modality (text and visual layout) of documents and show that the representations they provide are superior to pre-training on ImageNet, and subsequently better than all other self-training methods we have tested in the context of document image classification on Tobacco-3482 \cite{tobacco}.
   
   \item We make a qualitative analysis of the filters learned through our multi-modal pre-training method and show that they are similar to those learned through direct supervision, which makes our method a viable option for pre-training neural networks on document images.
   
\end{enumerate}

\section{Related Work}
\subsection{Transfer Learning}
One of the requirements of practicing statistical modeling is that the training and test data examples must be independent and identically distributed (i.i.d.). Transfer learning relaxes this hypothesis \cite{DBLP:journals/corr/abs-1808-01974}. In computer vision, most applications employ transfer learning through fine-tuning a model trained on the ImageNet Dataset \cite{imagenet_cvpr09}. Empirically, ImageNet models do transfer well on other subsequent tasks \cite{DBLP:journals/corr/abs-1805-08974,DBLP:journals/corr/HuhAE16}, even with little data for fine-tuning. However, it only has marginal performance gains for tasks in which labels are not well-represented in the ImageNet dataset. 

State-of-the-art results on related tasks such as object-detection \cite{DBLP:journals/corr/RenHG015} and instance segmentation \cite{DBLP:journals/corr/HeGDG17} are improved with the full ImageNet dataset used as pre-training, but data-efficient learning still remains a challenge.

\subsection{Self-Supervision}

Unsupervised learning methods for pre-training neural networks has sparked great interest in recent years. Given the large quantity of available data on the internet and the cost to rigorously annotate it, several methods have been proposed to learn general features. Most modern methods pre-train models to predict pseudo-labels on pretext tasks, to be fine-tuned on a supervised downstream task - usually with smaller amounts of data \cite{DBLP:journals/corr/abs-1902-06162}. 

With its roots in natural language processing, one of the most successful approaches is the skip-gram method \cite{NIPS2013_5021}, which provides general semantic textual representations by predicting the preceding and succeeding tokens from a single input token. More recent developments in natural language processing show promising results with models such as BERT \cite{radford2019language} and GPT-2 \cite{DBLP:journals/corr/abs-1810-04805}, which are pre-trained on a very large corpus of text to predict the next token. 

Similarly, this approach has been explored for images, with works trying to generate representations by "context prediction" \cite{DBLP:journals/corr/DoerschGE15}. Authors use a pretext task to classify the relative position of patches in an image. The same principle is used in works which explore solving jigsaw puzzles as a pretext task \cite{DBLP:journals/corr/NorooziF16,DBLP:journals/corr/CruzFCG17}. In both cases, the intuition is that a good performance on the patch classification task is directly correlated with a good performance on the downstream task, and with the network learning semantic information from the image.

Other self-supervision methods include predicting image rotations \cite{DBLP:journals/corr/abs-1803-07728}, image colorization \cite{DBLP:journals/corr/ZhangIE16} and even a multi-task model with several tasks at once \cite{DBLP:journals/corr/abs-1708-07860}. Furthermore, exemplar networks \cite{DBLP:journals/corr/DosovitskiySRB14} are trained to discriminate between a set of surrogate classes, to learn transformation invariant features. A more recent advancement in this area is Contrastive Predictive Coding \cite{DBLP:journals/corr/abs-1905-09272}, which is one of the most performing methods, for self-supervised pre-training.

An interesting multi-modal technique for self-supervision leverages a corpus of images from Wikipedia and their description \cite{DBLP:journals/corr/GomezPRKJ17}. The authors pre-train a network to predict the topic probabilities of the text description of an image, thereby leveraging the language context in which images appear. 

Clustering techniques have also been explored \cite{DBLP:journals/corr/abs-1810-02334,DBLP:journals/corr/abs-1805-00385,DBLP:journals/corr/abs-1905-01278} - by generating a classification pretext task with pseudo-labels based on cluster assignments. One method for unsupervised pre-training that makes very few assumptions about the input space is proposed by Bojanowski et al. \cite{Bojanowski:2017:ULP:3305381.3305435}. This approach trains a network to align a fixed set of target representations randomly sampled from the unit sphere.

Interestingly, a study by Kolesnikov et al. \cite{DBLP:journals/corr/abs-1901-09005} demonstrated that there is an inconsistency between self-supervision methods and network architectures. Some network architectures are better suited to encode image rotation, while others are better suited to handle patch information. We argue that this inconsistency also holds for datasets. These techniques show promising results on natural images, but very little research is devoted to learning good representations for document images, which have entirely different structural and semantic properties. 

\subsection{Document Analysis}
The representation of document images has a practical interest in commercial applications for tasks such as classification, retrieval, clustering, attribute extraction and historical document processing \cite{DBLP:journals/corr/abs-1804-10371}. Shallow features for representing documents \cite{DBLP:journals/corr/CsurkaLGA16} have proven to be less effective compared to deep features learned by a convolutional neural network. Several medium-scale datasets containing labeled document images are available, the ones used in this work being RVL-CDIP \cite{harley2015icdar} and Tobacco-3482 \cite{tobacco}.
For classification problems on document images, state-of-the-art approaches leverage domain knowledge of documents \cite{DBLP:journals/corr/AfzalKAL17}, combining features from the header, the footer and the contents of an image. Layout-methods are used in other works \cite{DBLP:journals/corr/abs-1809-08799,DBLP:journals/corr/YangYAKKG17,DBLP:journals/corr/abs-1907-06370} to make use of both textual information and their visual position in the image for use in extracting semantic structure.

One study by K{\"{o}}lsch et al. \cite{DBLP:journals/corr/abs-1711-05862} showed that pre-training networks using the RVL-CDIP dataset is better than pre-training with ImageNet in a supervised classification problem on the Tobacco-3482 dataset. Still, training from scratch is far worse than with ImageNet pre-training \cite{DBLP:journals/corr/abs-1905-09113}.

\section{Methods}

For our experiments, we implemented several methods for self-supervision and evaluated their performance on Tobacco-3482 document image classification task, where there is a limited amount of data. We implemented two Context-Free Networks (CFN), relative patch classification \cite{DBLP:journals/corr/DoerschGE15}, and solving jigsaw puzzles \cite{DBLP:journals/corr/NorooziF16}, which are patch-based, and, by design, are not using the broader context of the document. We also trained a model to predict image rotations \cite{DBLP:journals/corr/abs-1803-07728}, as a method that could intuitively make use of the layout and an input-agnostic method developed by Bojanowski et al. \cite{Bojanowski:2017:ULP:3305381.3305435}, which forces the model to learn mappings to the input image to noise vectors that are progressively aligned with deep features. We propose variations to context-free solving of jigsaw puzzles and to rotation prediction, which improves performance. We propose Jigsaw Whole, which is a pretext task to solve jigsaw puzzles, but with the whole image given as input, and predicting flips, which is in the same spirit of predicting rotations, but better suited for document images. 

We also developed a method that makes use of the information-rich textual modality: the model is tasked to predict the topic probabilities of the text present in the document using only the image as an input. This method is superior to ImageNet pre-training.

\subsection{Implementation Details}
 
Given the extensive survey of \cite{tensmeyer2017analysis}, we used the document images in grayscale format, resized to a fixed size of 384 x 384. Images are scaled so that the pixels fall in the interval $(-0.5, 0.5)$, by dividing by 255 and subtracting 0.5 \cite{6977258}.

Shear transformations or crops are usually used to improve the performance and robustness of CNNs on document images \cite{tensmeyer2017analysis}. We intentionally don't use augmentations during training or evaluation to speed up the process and lower the experiment's complexity.

InceptionV3 \cite{DBLP:journals/corr/SzegedyVISW15} architecture was used in all our experiments because of its popularity, performance and availability in common deep learning frameworks.

\subsection{Jigsaw Puzzles}

In the original paper for pre-training with solving Jigsaw puzzles \cite{DBLP:journals/corr/NorooziF16}, the authors propose a Context-Free Network architecture, with nine inputs, each being a crop from the original image. There, the pretext task is to reassemble the crops into the original image by predicting the permutation. 

This is sensible for natural images, which are invariant to zooms: objects appear at different scales in images, and random crops could contain information that is useful when fine-tuning on the main task. 

On the other hand, document images are more rigid in their structure. A random crop could end up falling in an area with blank space, or in the middle of a paragraph. Such crops contain no information for the layout of the document, and their relationship is not clear when processed independently. Moreover, text size changes relative to the crop size. As such, when fine-tuning, the text size is significantly smaller relative to the input size, which is inconsistent with the pretext task. 

\begin{figure}[!ht]
  \centering
  \begin{tikzpicture}
    \draw [rounded corners,fill=gray!20] (-5.5,0) -- (-1.5,0) -- (-1.5,4) -- (-5.5,4) -- cycle;
    \draw [rounded corners,fill=gray!20] (1.5,0) -- (5.5,0) -- (5.5,4) -- (1.5,4) -- cycle;
    \node[inner sep=0pt] (rvl) at (-3.5,2)
    {\includegraphics[width=.31\textwidth, height=.2\textheight]{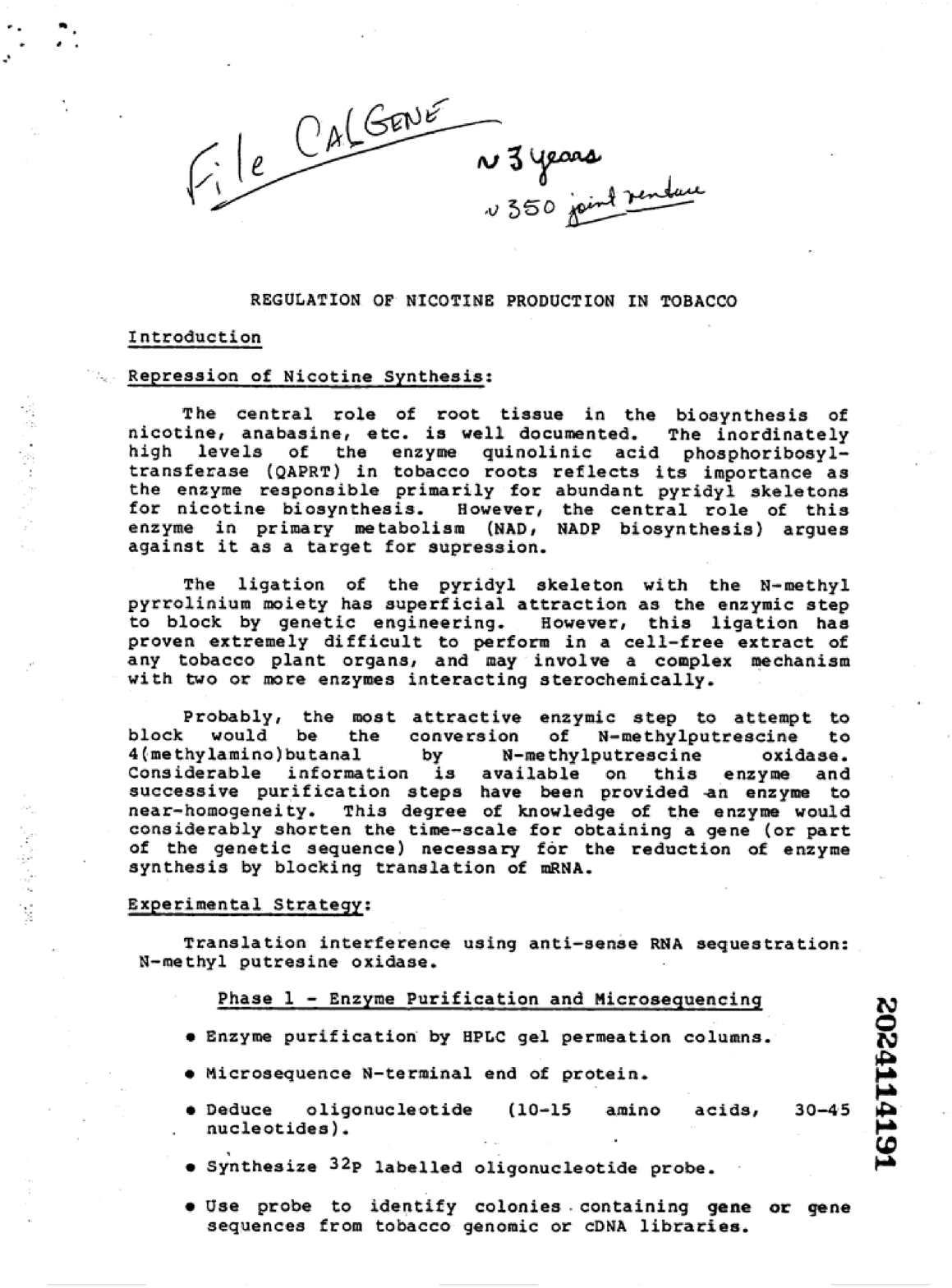}};
    \node[inner sep=0pt] (rvl) at (3.5,2)
    {\includegraphics[width=.31\textwidth, height=.2\textheight]{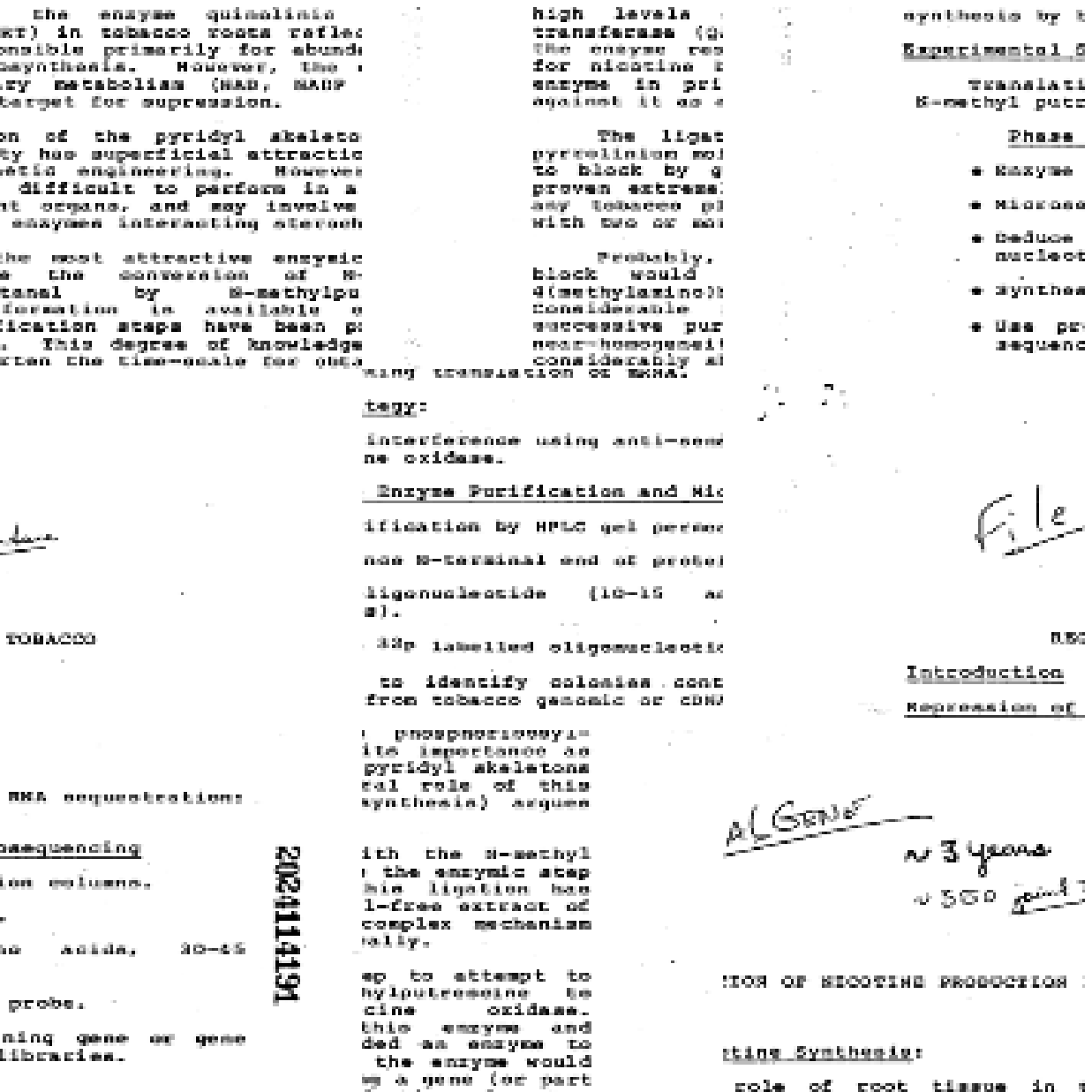}};
    \draw [->, line width=1, color=gray!100] (-1,2) .. controls (-1, 2) and (-1, 2) .. (1, 2);
  \end{tikzpicture}
  \caption{Input image for our Jigsaw Whole method. The scrambled image is given as is to a single network, without using siamese branches, such that context and layout information is preserved.}
  \label{fig:jigsaw}
\end{figure}

We propose a new way of pre-training by solving jigsaw puzzles with convolutional networks, by keeping the layout of the document visible to the model. After splitting the image into nine crops and shuffling them, we reassemble them into a single puzzle image. An example of the model input is exemplified in Figure \ref{fig:jigsaw}. We name this variation \textbf{Jigsaw Whole}. The intuition is that the convolutional network will better learn semantic features by leveraging the context of the document as a whole.

In order to obtain the final resolution, we resized the initial image at 384 x 384 pixels and then split it into nine crops of 128 x 128 pixels each. Using jitter (as recommended by Noroozi et al. \cite{DBLP:journals/corr/NorooziF16}) of 10 pixels, results in a resolution of 118 x 118 pixels for each of the nine patches. 

As described by Noroozi et al. \cite{DBLP:journals/corr/NorooziF16}, we chose only 100 out of 9! = 362880 possible permutations of the nine crops. Those were selected using a greedy algorithm to approximate the maximal average hamming distance between pairs of permutations from a set. 

\subsection{Relative Position of Patches}

Using a similar 3 x 3 grid as in the previous method, and based on Doersch et al. \cite{DBLP:journals/corr/DoerschGE15}, we implemented a siamese network to predict which is the position of a patch relative to the square in the center. The model has two inputs (the crop in the middle and one of the crops around it), and after the Global Average Pooling layer from InceptionV3, we added a fully-connected layer with 512 neurons, activated with a rectified linear unit and then a final fully-connected layer with eight neurons and softmax activation. For fine-tuning, we kept the representations created after the Global Average Pooling layer, ignoring the added fully-connected layer. To train the siamese network, we resized all the images to 384 x 384 pixels, then we created a grid with nine squares of 128 x 128 pixels each, and using jitter of 10 pixels, we obtained the input crops of 118 x 118.

Note that jitter is used both in solving jigsaw puzzles and predicting the relative position of patches, in order to prevent the network from solving the problem immediately by analysing the margins of the crops only (in which case it does not need to learn any other structural or semantic features).

Similar to solving jigsaw puzzles, predicting the relative position of patches suffer from the same problems of having too little context in a patch. We show that these methods perform poorly on document image classification.

\subsection{Rotations and Flips}

A recent method for self-supervision proposed by Gidaris et al. \cite{DBLP:journals/corr/abs-1803-07728} is the simple task of predicting image rotations. This task works for natural images quite well, since objects have an implicit orientation, and determining the rotation requires semantic information about that object. 
Documents, on the other hand, have only one orientation - upright. We pre-train our network to discriminate between 4 different orientations (0$^{\circ}$, 90$^{\circ}$, 180$^{\circ}$ and 270$^{\circ}$). It is evident that discriminating between (0$^{\circ}$, 180$^{\circ}$) pair and (90$^{\circ}$, 270$^{\circ}$) pair is trivial as the text lines are positioned differently. We argue that this is a shortcut for the model, and in this case, the task is not useful for learning semantic or layout information. 

Instead, we propose a new method, in the same spirit, by creating a pretext task that requires the model to discriminate between different flips of the document image. This way, the more challenging scenarios from the rotations methods are kept (in which text lines are always horizontal), and we argue that this forces the model to learn layout information or more fine-grained text features in order to discriminate between flips. It is worth noting that this method does not work in the case of natural images, as they are invariant to flips, at least across the vertical axis. In our experiments, we named this variation \textbf{Flips}.

\subsection{Multi-Modal Self-Supervised Pre-Training}

While plain computer vision methods are used with some degree of success, many applications do require textual information to be extracted from the documents. Be it the semantic structure of documents \cite{DBLP:journals/corr/YangYAKKG17}, or extracting attributes from financial documents \cite{DBLP:journals/corr/abs-1809-08799} or table understanding \cite{liu-etal-2019-graph,DBLP:journals/corr/abs-1904-12577,DBLP:journals/corr/abs-1905-13391}, the text modality present in documents is a rich source of information that can be leveraged to obtain better document representations. We assume that the visual document structure is correlated with the textual information present in the document. Audebert et al. \cite{DBLP:journals/corr/abs-1907-06370} use textual information to jointly classify documents from the RVL-CDIP dataset with significant results. Instead of jointly classifying, we explore self-supervised representation learning using text modality. 

The text is extracted by an OCR engine \cite{33418} making resulting text very noisy - many words have low document frequency due to OCR mistakes. While this should not be a problem given the large amount of data in the RVL-CDIP dataset, we do clean the text by lower-casing it, replacing all numbers with a single token and discarding any non-alpha-numeric characters.

\subsubsection{Text Topic Spaces}

Using textual modality to self-train a neural network was used by Gomez et al. \cite{DBLP:journals/corr/GomezPRKJ17} by exploiting the semantic context present in illustrated Wikipedia articles. The authors use the topic probabilities in the text as soft labels for the images in the article. Our approach is similar - we extract text from the RVL-CDIP dataset and analyse it using Latent Dirichlet Allocation \cite{Blei:2003:LDA:944919.944937} to extract topics. The CNN is then trained to predict the topic distribution, given only the image of the document. Different from the approach proposed by Gomez et al. \cite{DBLP:journals/corr/GomezPRKJ17}, there is a more intimate and direct correspondence between the document layout and its text content.

\begin{figure}[h!]
\centering
\includegraphics[width=0.8\textwidth]{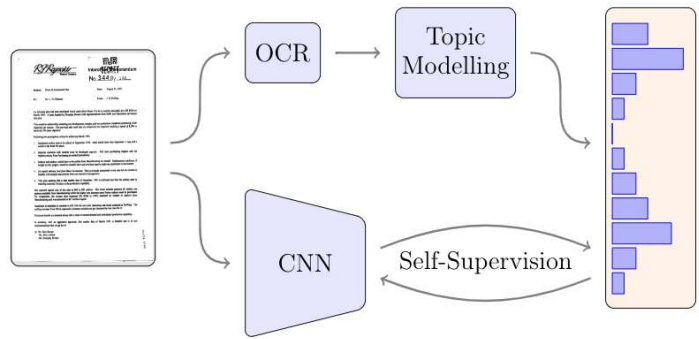}

\caption{General methodology for multi-modal self-supervision. The text from documents is extracted using an OCR engine, and then each of our method for topic modelling are used to generate topic probabilities. The neural network is then tasked to predict the topic probabilities using only the document image.} 
\label{fig:multi-modal}
\end{figure}

In this topic modeling method, we used soft-labels, as it was shown to improve performance in the context of knowledge distillation \cite{44873}. Soft-labels are also robust against noise \cite{10.1007/978-3-540-85563-7_14}, and have shown to increase performance for large-scale semi-supervised learning \cite{DBLP:journals/corr/abs-1905-00546}. Figure \ref{fig:multi-modal} depicts the general overview this method. Our intuition is that documents that are similar in topic spaces should also be similar in appearance. In our experiments, we named this self-supervision method \textbf{LDA Topic Spaces}.

\section{Experiments}
For the pre-training phase of the self-supervised methods, we used the training set from RVL-CDIP \cite{harley2015icdar}. RVL-CDIP is a dataset consisting of $400.000$ grayscale document images, of which $320.000$ are provided for training, $40.000$ for validation, and the remaining $40.000$ for testing. The images are labeled into 16 classes, some of which are also present in Tobacco-3482. Naturally, during our self-supervised pre-training experiments, we discard the labels. 
During the evaluation, we used the pre-trained models as feature extractors, and compute feature vectors for each image. We then trained a logistic classifier using L-BFGS \cite{Liu1989}. As shown by Kolesnikov et al. \cite{DBLP:journals/corr/abs-1901-09005}, a linear model is sufficient for evaluating the quality of features. Images used in the extraction phase come from the Tobacco-3482 dataset and are pre-processed exactly as during training. For partitioning the dataset, we used the same method as in \cite{DBLP:journals/corr/abs-1711-05862,DBLP:journals/corr/AfzalKAL17,6977258,harley2015icdar} for consistency and for a fair comparison with other works. We used Top-1 Accuracy as a metric, and we trained on a total of 10 to 100 images per class (with ten images increment), randomly sampled from Tobacco-3482. Testing was done on the rest of the images. We ran each experiment 10 times to reduce the odds of having a favourable configuration of training samples. Our evaluation scheme is designed for testing the performance in a document image classification setting with a limited amount of data.

In the particular case of LDA, we varied the number of topics and trained three models tasked to predict the topic probabilities of 16, 32, and 64 topics. By using this form of soft-clustering in the topic space, the model benefited from having a finer-grained topic distribution.

In our experiments, we used two supervised benchmarks: a model pre-trained on ImageNet and a model pre-trained on RVL-CDIP. The supervised pre-training methods have an obvious advantage, due to the high amount of consistent and correct information present in annotations. Consistent with other works \cite{DBLP:journals/corr/abs-1711-05862}, supervised RVL-CDIP pre-training is far superior.

\begin{figure}[h!]
  \centering
  \includegraphics[width=0.9\textwidth]{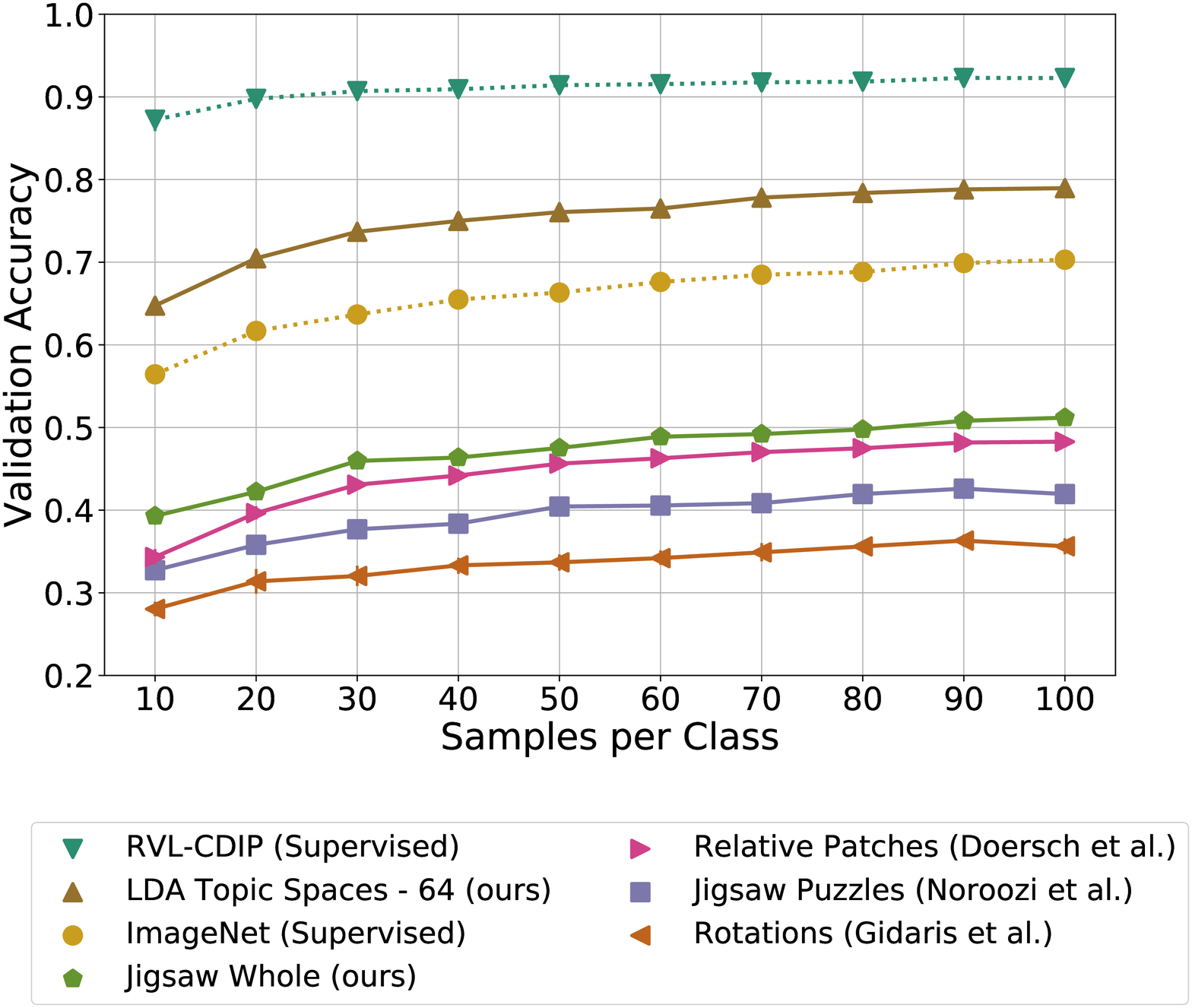}
  \caption{Performance on fine-tuning for some of the more relevant methods on different sample sizes from Tobacco-3482. Our proposed methods have higher accuracy than previous attempts on natural images. Multi-modal self-supervision using LDA (LDA Topic Spaces) with 64 topics is significantly higher than supervised ImageNet pre-training and much higher than the other self-supervised methods we tested.}
  \label{fig:data-efficient}
\end{figure}

\section{Results}
In Figure \ref{fig:data-efficient}, we show some of the more relevant methods in the evaluation scheme. Supervised RVL-CDIP is, unsurprisingly, the most performing method, and our self-supervised multi-modal approach has a significantly higher accuracy overall when compared to supervised ImageNet pre-training. Features extracted from patch-based methods and methods which rely only on layout information are not discriminative enough to have higher accuracy. This is also consistent with the original works \cite{DBLP:journals/corr/NorooziF16,DBLP:journals/corr/DoerschGE15,Bojanowski:2017:ULP:3305381.3305435} in which self-supervised pre-training did not provide a boost in performance compared to the supervised baseline.

Relative Patches and Jigsaw Puzzles have only modest performance. Both these methods initially used a "context-free" approach. Our variation of Jigsaw Puzzles - Jigsaw Whole - works around this by actually including more context. Receiving the entire document helps the network to learn features that are relevant for the layout. Features learned this way are more discriminative for the classes in Tobacco-3842. In the case of Relative Patches, there is no sensible way to include more context in the input, as stitching together two patches changes the aspect ratio of the input. 

In Table \ref{table:results}, we present the mean accuracy for 100 samples per class on all methods. We also implemented the work of Bojanowski et al. \cite{Bojanowski:2017:ULP:3305381.3305435}, to pre-train a model by predicting noise. This method is very general and assumes very little of the inputs. We discovered that it was better than predicting rotations. The features extracted by predicting rotations were, in fact, the weakest, as this task is far too easy for a model in the case of document images. Our variation, predicting flips, provides a much harder task, which translates into a better performance on the classification task.

\begin{table}[h!]
\centering
 \begin{tabular*}{0.7\textwidth}{l @{\extracolsep{\fill}} r} 
 \toprule
 \textbf{Method} & \textbf{Accuracy} \%\\
 \midrule
Rotations (Gidaris et al.) & 35.63 $\pm$ 2.74 \\
Noise as Targets (Bojanowski et al.) & 40.85 $\pm$ 1.58 \\
Jigsaw Puzzles (Noroozi et al.) & 41.95 $\pm$ 1.71 \\
Relative Patches (Doersch et al.) & 48.28 $\pm$ 1.69 \\
\textbf{Flips (ours)} & \textbf{50.52} \textbf{$\pm$} \textbf{1.93} \\
\textbf{Jigsaw Whole (ours)} & \textbf{51.19} \textbf{$\pm$} \textbf{2.06} \\
ImageNet pre-training (Supervised) & 70.3 $\pm$ 1.5 \\
LDA Topic Spaces - 16 (ours) & 75.75 $\pm$ 1.47 \\
LDA Topic Spaces - 32 (ours) & 76.22 $\pm$ 1.72 \\
\textbf{LDA Topic Spaces - 64 (ours)} & \textbf{78.96} \textbf{$\pm$} \textbf{1.42} \\
RVL-CDIP pre-training (Supervised) & 92.28 $\pm$ 0.8 \\
\bottomrule
\label{table:results}
\end{tabular*}
\caption{Results on our pre-training experiments. We also implemented Noise as Targets (NAT) \cite{Bojanowski:2017:ULP:3305381.3305435} as an input-agnostic method for self-supervision. For LDA, we tried multiple number of topics and have decided upon 64 before experiencing diminishing returns. All presented methods employ self-supervised pre-training, unless otherwise specified.}
\end{table}

In the case of topic modeling, we argue that the boost in performance is due to the high correlation between the similarity in the topic space and the similarity in the "image" space. The 16 classes in RVL-CDIP are sufficiently dissimilar between them in terms of topics (i.e., news article, resume, advertisement), and each class of documents has a specific layout. Surprisingly, LDA with 16 topics was the weakest. A finer-grained topic distribution helped the model in learning more discriminative features.

\subsection{Qualitative Analysis}
\begin{figure}[h!]
\centering
    \includegraphics[width=\textwidth]{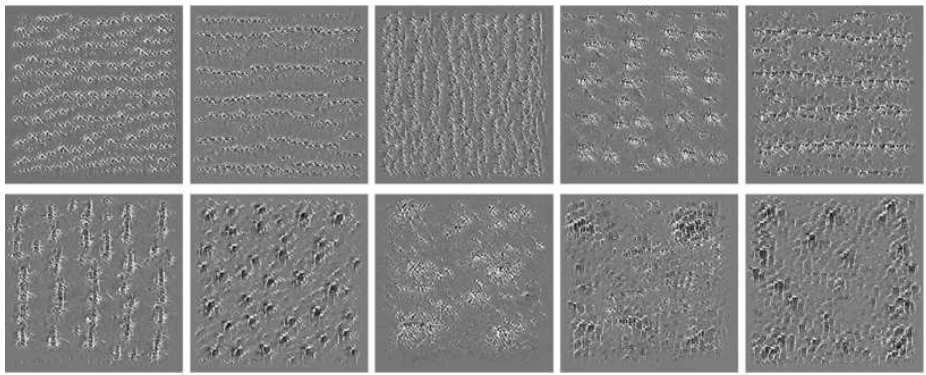}

  \caption{Gradient ascent visualization of filters learned through supervised RVL pre-training. Patterns for text lines, columns and paragraphs appear. }
  \label{fig:rvl-filters}
\end{figure}

\begin{figure}[h!]
\centering
    \includegraphics[width=\textwidth]{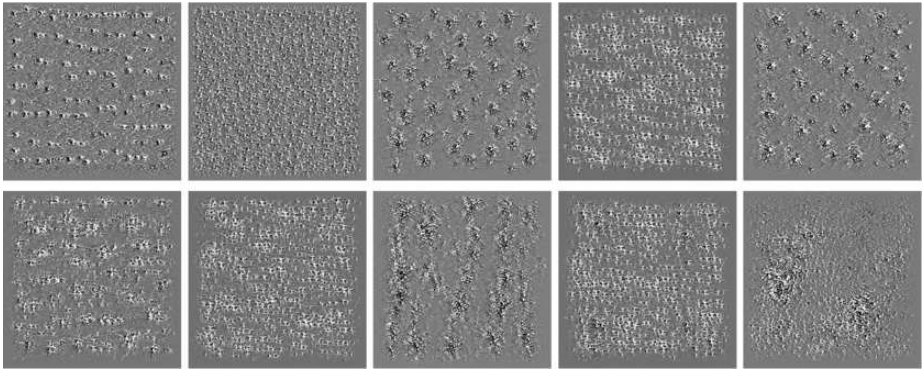}
  \caption{Gradient ascent visualization of filters learned by LDA pre-training. Activation patterns that correspond to words and paragraphs emerge. More distinctively, patterns that appear to resemble words are more frequent than in the supervised setting.}
  \label{fig:lda-filters}
\end{figure}

For the qualitative analysis, we compare filters learned through LDA self - supervision with those learned through RVL-CDIP supervised pre-training. In Figures \ref{fig:rvl-filters} and \ref{fig:lda-filters}, we show gradient ascent visualizations learned by both methods - a randomly initialized input image is progressively modified such that the mean activation of a given filter is maximized. The filters shown are from increasing depths in the InceptionV3 network, from conv2d\_10 to conv2d\_60.

In Figure \ref{fig:rvl-filters} there are clear emerging patters that correspond to text lines, columns and paragraphs - general features that apply to a large subset of documents.  

In Figure \ref{fig:lda-filters}, we show filters learned by LDA self - supervision. In contrast to the "gold-standard" filter learned by RVL-CDIP supervision, the patterns that emerge here are frequently more akin to what appears to be words. This is a direct consequence of the way LDA constructs topics, in which a topic is based on a bag-of-words model. Our neural network, therefore, has a high response in regions corresponding to particular words in the image. The features learned this way are nonetheless discriminative for document images, as patterns for paragraphs and text columns still emerge. Another particularity of these filters is that they are noisier than those learned by direct supervision. This likely results from the soft and noisy labels generated by the topic model, and due to imperfect text extracted by the OCR engine. Naturally, features learned from ImageNet pre-training are qualitatively different and more general from filters that are specialized in extracting information from document images. See Olah et al. \cite{olah2017feature} for a comprehensive visualization of InceptionV3 trained on ImageNet.

\section{Conclusions}
We have explored self-supervision methods that were previously introduced in the realm of natural images and showed that document images have a more rigid visual structure, which makes patch-based methods not work as well on them. To that end, we proposed slight alterations that exploit documents' visual properties: self-supervised pre-training by predicting flips and by solving jigsaw puzzles with the whole layout present in the input. 

Documents are inherently multi-modal. As such, by extracting text from the document images, we developed a method to pre-train a network, in a self-supervised manner, to predict topics generated by Latent Dirichlet Allocation. This method outperforms the strong baseline of supervised pre-training from ImageNet. We also show that the features learned this way are closely related to those learned through direct supervision on RVL-CDIP, making this method a viable method for pre-training neural networks on document images.

\section{Acknowledgements}
We want to express our appreciation for everyone involved at Sparktech Software, for fruitful discussions, and much-needed suggestions. In particular, we would like to thank Andrei Manea, Andrei Sb\^{a}rcea, Antonio B\u{a}rb\u{a}l\u{a}u, and Andrei Iu\cb{s}an.

\bibliographystyle{splncs04}
\bibliography{refs.bib}

\begin{thebibliography}{10}
\providecommand{\url}[1]{\texttt{#1}}
\providecommand{\urlprefix}{URL }
\providecommand{\doi}[1]{https://doi.org/#1}

\bibitem{DBLP:journals/corr/AfzalKAL17}
Afzal, M.Z., K{\"{o}}lsch, A., Ahmed, S., Liwicki, M.: Cutting the error by
  half: Investigation of very deep {CNN} and advanced training strategies for
  document image classification. CoRR  \textbf{abs/1704.03557} (2017),
  \url{http://arxiv.org/abs/1704.03557}

\bibitem{DBLP:journals/corr/abs-1907-06370}
Audebert, N., Herold, C., Slimani, K., Vidal, C.: Multimodal deep networks for
  text and image-based document classification. CoRR  \textbf{abs/1907.06370}
  (2019), \url{http://arxiv.org/abs/1907.06370}

\bibitem{DBLP:journals/corr/GomezPRKJ17}
i~Bigorda, L.G., Patel, Y., Rusi{\~{n}}ol, M., Karatzas, D., Jawahar, C.V.:
  Self-supervised learning of visual features through embedding images into
  text topic spaces. CoRR  \textbf{abs/1705.08631} (2017),
  \url{http://arxiv.org/abs/1705.08631}

\bibitem{Blei:2003:LDA:944919.944937}
Blei, D.M., Ng, A.Y., Jordan, M.I.: Latent dirichlet allocation. J. Mach.
  Learn. Res.  \textbf{3},  993--1022 (Mar 2003),
  \url{http://dl.acm.org/citation.cfm?id=944919.944937}

\bibitem{Bojanowski:2017:ULP:3305381.3305435}
Bojanowski, P., Joulin, A.: Unsupervised learning by predicting noise. In:
  Proceedings of the 34th International Conference on Machine Learning - Volume
  70. pp. 517--526. ICML'17, JMLR.org (2017),
  \url{http://dl.acm.org/citation.cfm?id=3305381.3305435}

\bibitem{DBLP:journals/corr/abs-1905-01278}
Caron, M., Bojanowski, P., Mairal, J., Joulin, A.: Leveraging large-scale
  uncurated data for unsupervised pre-training of visual features. CoRR
  \textbf{abs/1905.01278} (2019), \url{http://arxiv.org/abs/1905.01278}

\bibitem{DBLP:journals/corr/CruzFCG17}
Cruz, R.S., Fernando, B., Cherian, A., Gould, S.: Deeppermnet: Visual
  permutation learning. CoRR  \textbf{abs/1704.02729} (2017),
  \url{http://arxiv.org/abs/1704.02729}

\bibitem{DBLP:journals/corr/CsurkaLGA16}
Csurka, G., Larlus, D., Gordo, A., Almaz{\'{a}}n, J.: What is the right way to
  represent document images? CoRR  \textbf{abs/1603.01076} (2016),
  \url{http://arxiv.org/abs/1603.01076}

\bibitem{imagenet_cvpr09}
Deng, J., Dong, W., Socher, R., Li, L.J., Li, K., Fei-Fei, L.: {ImageNet: A
  Large-Scale Hierarchical Image Database}. In: CVPR09 (2009)

\bibitem{DBLP:journals/corr/abs-1810-04805}
Devlin, J., Chang, M., Lee, K., Toutanova, K.: {BERT:} pre-training of deep
  bidirectional transformers for language understanding. CoRR
  \textbf{abs/1810.04805} (2018), \url{http://arxiv.org/abs/1810.04805}

\bibitem{DBLP:journals/corr/DoerschGE15}
Doersch, C., Gupta, A., Efros, A.A.: Unsupervised visual representation
  learning by context prediction. CoRR  \textbf{abs/1505.05192} (2015),
  \url{http://arxiv.org/abs/1505.05192}

\bibitem{DBLP:journals/corr/abs-1708-07860}
Doersch, C., Zisserman, A.: Multi-task self-supervised visual learning. CoRR
  \textbf{abs/1708.07860} (2017), \url{http://arxiv.org/abs/1708.07860}

\bibitem{DBLP:journals/corr/DosovitskiySRB14}
Dosovitskiy, A., Springenberg, J.T., Riedmiller, M.A., Brox, T.: Discriminative
  unsupervised feature learning with convolutional neural networks. CoRR
  \textbf{abs/1406.6909} (2014), \url{http://arxiv.org/abs/1406.6909}

\bibitem{DBLP:journals/corr/abs-1803-07728}
Gidaris, S., Singh, P., Komodakis, N.: Unsupervised representation learning by
  predicting image rotations. CoRR  \textbf{abs/1803.07728} (2018),
  \url{http://arxiv.org/abs/1803.07728}

\bibitem{harley2015icdar}
Harley, A.W., Ufkes, A., Derpanis, K.G.: Evaluation of deep convolutional nets
  for document image classification and retrieval. In: International Conference
  on Document Analysis and Recognition ({ICDAR}) (2015)

\bibitem{DBLP:journals/corr/HeGDG17}
He, K., Gkioxari, G., Doll{\'{a}}r, P., Girshick, R.B.: Mask {R-CNN}. CoRR
  \textbf{abs/1703.06870} (2017), \url{http://arxiv.org/abs/1703.06870}

\bibitem{DBLP:journals/corr/abs-1905-09272}
H{\'{e}}naff, O.J., Razavi, A., Doersch, C., Eslami, S.M.A., van~den Oord, A.:
  Data-efficient image recognition with contrastive predictive coding. CoRR
  \textbf{abs/1905.09272} (2019), \url{http://arxiv.org/abs/1905.09272}

\bibitem{44873}
Hinton, G., Vinyals, O., Dean, J.: Distilling the knowledge in a neural
  network. In: NIPS Deep Learning and Representation Learning Workshop (2015),
  \url{http://arxiv.org/abs/1503.02531}

\bibitem{DBLP:journals/corr/abs-1904-12577}
Holecek, M., Hoskovec, A., Baudis, P., Klinger, P.: Line-items and table
  understanding in structured documents. CoRR  \textbf{abs/1904.12577} (2019),
  \url{http://arxiv.org/abs/1904.12577}

\bibitem{DBLP:journals/corr/abs-1810-02334}
Hsu, K., Levine, S., Finn, C.: Unsupervised learning via meta-learning. CoRR
  \textbf{abs/1810.02334} (2018), \url{http://arxiv.org/abs/1810.02334}

\bibitem{DBLP:journals/corr/HuhAE16}
Huh, M., Agrawal, P., Efros, A.A.: What makes imagenet good for transfer
  learning? CoRR  \textbf{abs/1608.08614} (2016),
  \url{http://arxiv.org/abs/1608.08614}

\bibitem{DBLP:journals/corr/abs-1902-06162}
Jing, L., Tian, Y.: Self-supervised visual feature learning with deep neural
  networks: {A} survey. CoRR  \textbf{abs/1902.06162} (2019),
  \url{http://arxiv.org/abs/1902.06162}

\bibitem{6977258}
{Kang}, L., {Kumar}, J., {Ye}, P., {Li}, Y., {Doermann}, D.: Convolutional
  neural networks for document image classification. In: 2014 22nd
  International Conference on Pattern Recognition. pp. 3168--3172 (Aug 2014).
  \doi{10.1109/ICPR.2014.546}

\bibitem{DBLP:journals/corr/abs-1809-08799}
Katti, A.R., Reisswig, C., Guder, C., Brarda, S., Bickel, S., H{\"{o}}hne, J.,
  Faddoul, J.B.: Chargrid: Towards understanding 2d documents. CoRR
  \textbf{abs/1809.08799} (2018), \url{http://arxiv.org/abs/1809.08799}

\bibitem{DBLP:journals/corr/abs-1901-09005}
Kolesnikov, A., Zhai, X., Beyer, L.: Revisiting self-supervised visual
  representation learning. CoRR  \textbf{abs/1901.09005} (2019),
  \url{http://arxiv.org/abs/1901.09005}

\bibitem{DBLP:journals/corr/abs-1711-05862}
K{\"{o}}lsch, A., Afzal, M.Z., Ebbecke, M., Liwicki, M.: Real-time document
  image classification using deep {CNN} and extreme learning machines. CoRR
  \textbf{abs/1711.05862} (2017), \url{http://arxiv.org/abs/1711.05862}

\bibitem{DBLP:journals/corr/abs-1805-08974}
Kornblith, S., Shlens, J., Le, Q.V.: Do better imagenet models transfer better?
  CoRR  \textbf{abs/1805.08974} (2018), \url{http://arxiv.org/abs/1805.08974}

\bibitem{tobacco}
Kumar, J., Ye, P., Doermann, D.: Structural similarity for document image
  classification and retrieval. Pattern Recognition Letters  \textbf{43},
  119–126 (07 2014). \doi{10.1016/j.patrec.2013.10.030}

\bibitem{li2019tablebank}
Li, M., Cui, L., Huang, S., Wei, F., Zhou, M., Li, Z.: Tablebank: Table
  benchmark for image-based table detection and recognition. arXiv preprint
  arXiv:1903.01949  (2019)

\bibitem{Liu1989}
Liu, D.C., Nocedal, J.: On the limited memory bfgs method for large scale
  optimization. Mathematical Programming  \textbf{45}(1),  503--528 (Aug 1989).
  \doi{10.1007/BF01589116}, \url{https://doi.org/10.1007/BF01589116}

\bibitem{liu-etal-2019-graph}
Liu, X., Gao, F., Zhang, Q., Zhao, H.: Graph convolution for multimodal
  information extraction from visually rich documents. In: Proceedings of the
  2019 Conference of the North {A}merican Chapter of the Association for
  Computational Linguistics: Human Language Technologies, Volume 2 (Industry
  Papers). pp. 32--39. Association for Computational Linguistics, Minneapolis,
  Minnesota (Jun 2019). \doi{10.18653/v1/N19-2005},
  \url{https://www.aclweb.org/anthology/N19-2005}

\bibitem{NIPS2013_5021}
Mikolov, T., Sutskever, I., Chen, K., Corrado, G., Dean, J.: Distributed
  representations of words and phrases and their compositionality. vol.
  abs/1310.4546 (2013), \url{http://arxiv.org/abs/1310.4546}

\bibitem{DBLP:journals/corr/NorooziF16}
Noroozi, M., Favaro, P.: Unsupervised learning of visual representations by
  solving jigsaw puzzles. CoRR  \textbf{abs/1603.09246} (2016),
  \url{http://arxiv.org/abs/1603.09246}

\bibitem{DBLP:journals/corr/abs-1805-00385}
Noroozi, M., Vinjimoor, A., Favaro, P., Pirsiavash, H.: Boosting
  self-supervised learning via knowledge transfer. CoRR
  \textbf{abs/1805.00385} (2018), \url{http://arxiv.org/abs/1805.00385}

\bibitem{olah2017feature}
Olah, C., Mordvintsev, A., Schubert, L.: Feature visualization. Distill
  (2017). \doi{10.23915/distill.00007},
  https://distill.pub/2017/feature-visualization

\bibitem{DBLP:journals/corr/abs-1804-10371}
Oliveira, S.A., Seguin, B., Kaplan, F.: dhsegment: {A} generic deep-learning
  approach for document segmentation. CoRR  \textbf{abs/1804.10371} (2018),
  \url{http://arxiv.org/abs/1804.10371}

\bibitem{DBLP:journals/corr/abs-1905-13391}
Qasim, S.R., Mahmood, H., Shafait, F.: Rethinking table parsing using graph
  neural networks. CoRR  \textbf{abs/1905.13391} (2019),
  \url{http://arxiv.org/abs/1905.13391}

\bibitem{radford2019language}
Radford, A., Wu, J., Child, R., Luan, D., Amodei, D., Sutskever, I.: Language
  models are unsupervised multitask learners  (2019)

\bibitem{DBLP:journals/corr/RenHG015}
Ren, S., He, K., Girshick, R.B., Sun, J.: Faster {R-CNN:} towards real-time
  object detection with region proposal networks. CoRR  \textbf{abs/1506.01497}
  (2015), \url{http://arxiv.org/abs/1506.01497}

\bibitem{33418}
Smith, R.: An overview of the tesseract ocr engine. In: Proc. Ninth Int.
  Conference on Document Analysis and Recognition (ICDAR). pp. 629--633 (2007)

\bibitem{DBLP:journals/corr/abs-1905-09113}
Studer, L., Alberti, M., Pondenkandath, V., Goktepe, P., Kolonko, T., Fischer,
  A., Liwicki, M., Ingold, R.: A comprehensive study of imagenet pre-training
  for historical document image analysis. CoRR  \textbf{abs/1905.09113} (2019),
  \url{http://arxiv.org/abs/1905.09113}

\bibitem{DBLP:journals/corr/SzegedyVISW15}
Szegedy, C., Vanhoucke, V., Ioffe, S., Shlens, J., Wojna, Z.: Rethinking the
  inception architecture for computer vision. CoRR  \textbf{abs/1512.00567}
  (2015), \url{http://arxiv.org/abs/1512.00567}

\bibitem{DBLP:journals/corr/abs-1808-01974}
Tan, C., Sun, F., Kong, T., Zhang, W., Yang, C., Liu, C.: A survey on deep
  transfer learning. CoRR  \textbf{abs/1808.01974} (2018),
  \url{http://arxiv.org/abs/1808.01974}

\bibitem{tensmeyer2017analysis}
Tensmeyer, C., Martinez, T.: Analysis of convolutional neural networks for
  document image classification (2017)

\bibitem{10.1007/978-3-540-85563-7_14}
Thiel, C.: Classification on soft labels is robust against label noise. In:
  Lovrek, I., Howlett, R.J., Jain, L.C. (eds.) Knowledge-Based Intelligent
  Information and Engineering Systems. pp. 65--73. Springer Berlin Heidelberg,
  Berlin, Heidelberg (2008)

\bibitem{DBLP:journals/corr/abs-1905-00546}
Yalniz, I.Z., J{\'{e}}gou, H., Chen, K., Paluri, M., Mahajan, D.: Billion-scale
  semi-supervised learning for image classification. CoRR
  \textbf{abs/1905.00546} (2019), \url{http://arxiv.org/abs/1905.00546}

\bibitem{DBLP:journals/corr/YangYAKKG17}
Yang, X., Y{\"{u}}mer, M.E., Asente, P., Kraley, M., Kifer, D., Giles, C.L.:
  Learning to extract semantic structure from documents using multimodal fully
  convolutional neural network. CoRR  \textbf{abs/1706.02337} (2017),
  \url{http://arxiv.org/abs/1706.02337}

\bibitem{DBLP:journals/corr/ZhangIE16}
Zhang, R., Isola, P., Efros, A.A.: Colorful image colorization. CoRR
  \textbf{abs/1603.08511} (2016), \url{http://arxiv.org/abs/1603.08511}

\bibitem{zhuang2019comprehensive}
Zhuang, F., Qi, Z., Duan, K., Xi, D., Zhu, Y., Zhu, H., Xiong, H., He, Q.: A
  comprehensive survey on transfer learning (2019)

\end{thebibliography}

\end{document}